\documentclass[letterpaper]{article}
\usepackage{booktabs}
\usepackage{times}
\usepackage{epsfig}
\usepackage{graphicx}
\usepackage{amsmath}
\usepackage{amssymb}
\usepackage{array}
\usepackage{multirow}
\usepackage{url}
\usepackage{wrapfig}
\usepackage{booktabs}

\usepackage{subfig}
\usepackage[pagebackref=true,breaklinks=true,letterpaper=true,colorlinks,bookmarks=false]{hyperref}
\begin{document}
\title{Unsupervised learning with sparse space-and-time autoencoders}

\author{
Benjamin Graham\\
Facebook AI Research\\
{\tt\small benjamingraham@fb.com}\\
}

\maketitle

\begin{abstract}
We use spatially-sparse two, three and four dimensional convolutional
autoencoder networks to model sparse structures in 2D space, 3D space,
and $3+1=4$ dimensional space-time. We evaluate the resulting latent
spaces by testing their usefulness for downstream tasks. Applications
are to handwriting recognition in 2D, segmentation for parts in 3D
objects, segmentation for objects in 3D scenes, and body-part segmentation
for 4D wire-frame models generated from motion capture data.
\end{abstract}

\section{Introduction}

Convolutional networks were initially developed for supervised learning.
They are used in deep learning to classify two-dimensional \emph{spatial}
information such as hand writing samples and photographs \cite{lecun-98}. In the one
dimensional setting, they have been applied to \emph{temporal} data
such as audio recordings of speech and music, and writing encoded
at either the character level or the word level. In the three dimensional
setting, applications have included medical scans, object detection
for self driving cars, and object recognition from RGB-D photos. Videos,
with their two spatial dimensions and one time dimension can also
be seen as $2+1=3$ dimensional objects for purposes of applying convolutional
networks \cite{journals/corr/TranBFTP14}. The movement of 3D objects happens in $3+1=4$ dimensional
space-time, but 4D ConvNets are relatively unexplored.

\subsection{Autoencoder networks and unsupervised learning}

Gathering labeled datasets is onerous, so unsupervised learning is
an important research area \cite{RUMELHART-HINTON-WILLIAMS86}. Autoencoder networks encode the input
into a latent space. They can be written in two parts,
\begin{align*}
\textrm{latent} & =\textrm{encoder}(\textrm{input}),\\
\textrm{output} & =\textrm{decoder}(\textrm{latent}),
\end{align*}
where the encoder and decoder networks are trained jointly to minimize
the distance between the input and the output for some training set.
This is called \emph{unsupervised} learning.
The latent space captures much of the information from the input,
and so it can be used for \emph{downstream} tasks. Convolutional autoencoder networks that combine downsampling and upsampling layers can be used to learn latent representation of spatial data \cite{zeiler10, AdaptiveDeconvolutional}.
Unsupervised learning has also been done with 2D ConvNets in other ways, such as
solving jigsaw puzzles made up of image fragments \cite{jigsaw}, and learning to
predict the identity of images within an large unlabeled database \cite{exemplar, predictingNoise, journals/corr/abs-1807-05520}.

In natural language processing unsupervised (or self-supervised) techniques
such as Word2Vec \cite{Word2Vec} and FastText \cite{FastTextClassification} have shown that features trained simply
to predict the environment are useful for a range of down-stream tasks,
such as question answering and machine translation, where they may
be fed into a convolutional or recurrent network as input features.
Word2Vec is a low-rank factorization
of the matrix of nearby word co-occurrences. The implicit language
model is to guess a missing word based on the immediate context.

For autoencoders trained on image datasets with the $L_{2}$ metric,
the output is typically blurry. For an autoencoder network to reconstruct
a sharp image of a furry animal, you need to capture the location
of every visible hair, even though the reconstruction would look fine
with the hairs arranged slightly differently. Overly precise information
about the location of fine details that form part of a larger pattern
is unlikely to be of any interest for downstream tasks. This problem
has driven research into variational autoencoders and GANs.

\subsection{Encoder-decoder networks for image to image transformations}

Convolutional networks combining encoder and decoder components can
also be used to perform image to image mapping operations such as
segmentation \cite{long2015fully} and image editing \cite{CycleGAN, FaderNetworks}. Downsampling can be performed
by max-pooling or strided convolution, and upsampling can be performed by
unpooling or transpose convolutions. Shortcut connections linking hidden layers at the same spatial scale in the encoder and decoder networks improve accuracy \cite{ronneberger2015unet}.

\subsection{Spatially-sparse input in $d\ge2$ dimensions}

The success of two-dimensional convolutional networks operating on
dense 2D images has spurred interest in higher dimensional machine learning.
Larger 3D datasets have been released recently, such as ShapeNet\footnote{\href{https://www.shapenet.org/}{https://www.shapenet.org/}} and ScanNet \footnote{\href{http://www.scan-net.org/}{http://www.scan-net.org/}}.
However, higher dimensional ConvNets have not yet becomes as widely used as their 2D counterparts.
Limiting factors have included:
\begin{itemize}
\item High computational overhead in terms of floating-point operations (FLOPs) and memory.
\item Restricted software support in popular machine learning software packages.
\end{itemize}
However, a flip side to the curse of dimensionality is that in higher dimensional settings, sparsity becomes more likely.
\begin{description}
\item[2D] A pen drawing the letter `z' in a $n\times n$ grid might visit
approximately $3n$ of the $n^{2}$ locations, suggesting that handwriting
might be 5-10\% occupied.
\item[3D] A bounding cuboid around the Eiffel Tower is 99.98\% empty
space (air) and just 0.02\% iron.
\item[4D] A space-time
path in a hyper-cube of size $64^{4}$ visits just 0.0004\%
of the lattice sites.
\end{description}
Sparse data can be represented using either point clouds or sparse tensors.
In the case of tensors, recall that $d$-dimensional ConvNets
typically operate on $d+1$ or $d+2$ dimensional dense tensors, the
extra dimensions represent the feature channels and possibly the
batch size; e.g. the input to an AlexNet 2D ConvNet will be a tensor of size
$3\times224\times224$ where 3 is the number of input channels of
an RGB image, and $224\times224$ is the spatial size.

For 3D and 4D objects, the most intuitive form of sparsity
is \emph{spatial/spatio-temporal sparsity}: each location in space or space-time is either:
\begin{description}
\item[active] in which case the value of each of the feature channels
at that location is typically non-zero, or
\item[inactive] with all of the feature channels taking value zero.
\end{description}
This regularity in the sparsity structure means
that the vectors of feature channels at active locations can be represented by contiguous tensors, which is important for efficient computation.
To exploit this spatial sparsity, a variety of sparse convolutional networks have been developed:
\begin{itemize}
\item Permutohedral lattice CNNs \cite{PermutohedralLatticeCNN} operate on set of points with floating-point coordinates. Convolutions are implemented by projecting onto a permutohedral lattice and back using splat-convolve-slice operations at each level of the network.
\item SparseConvNets \cite{graham2015sparse} are mathematically identical to regular dense ConvNets, but implemented using sparse data structures. A ground state hidden vector was used at each level of the network to describe the hidden vectors that could see no input. A drawback of SparseConvNets is that deep stacks of size-3 stride-1 convolutions \cite{VGG} quickly reduces the level of sparsity due to the blurring effect of convolutions.
\item OctNets\cite{riegler2016octnet} provided an alternative form of sparsity.
Empty portions of the input are amalgamated into dyadic cubes that then share a hidden state vector.
\item Vote3Deep \cite{engelcke2017vote3deep} uses dense tensors, but it is sparse computationally, and uses a loss function during training to promote sparsity.
\item Kd-Networks \cite{klokov2017escape} implements a type of graph convolution on the Kd-tree of point clouds.
\item PointNet \cite{qi2016pointnet} treats coordinates as floating point features for a fully connected neural network.
\end{itemize}
To allow computational resources to be more focused on the active regions, both SparseConvNets and OctNets have both been modified to remove the hidden vectors corresponding to empty regions:
\begin{itemize}
\item Submanifold SparseConvNets (SSCN) \cite{graham2017submanifoldB} discards the ground state hidden vector, and introduces a parsimonious convolution operation that is restricted to only operates on already active sites, hence eliminating the blurring problem.
\item Octree-based Convolutional Neural Networks (O-CNN) \cite{OcnnNoGroundState}
remove the hidden vectors for empty OctTree cells.
\end{itemize}
In both cases, stride-1 convolutions are performed in a sparsity preserving manner,
while stride-2 convolutions used for downsampling by a factor of $2\times$ are greedy.

In terms of implementation, at each layer SSCN uses a hash table to store the set of active locations; it can be compiled to support any positive integer dimension. O-CNN uses a hybrid mix of OctTrees and hash tables to store the spatial structure, so it is hard-coded to operate in three dimensions, but could in principle be extended to support other dimensions using $2^d$-trees.
The networks we introduce in the next section could be defined in either of the two frameworks. We use SSCN as it already supports dimensions 2, 3 and 4.

\section{Sparse Autoencoders}
\begin{figure*}
\begin{center}
\makebox[0pt][c]{%
\begin{minipage}{0.7\paperwidth}
\begin{minipage}[t]{0.55\textwidth}
\begin{tabular}{lccl}
\multicolumn{4}{c}{\sc Encoder}\\
\toprule
\bf Layer & \bf Input & \bf Output & \bf Sparsity\\
\midrule
SSC$(k,k,3)$ & $k\times$16$^{d}$ & $k\times$16$^{d}$ & a\\
SC$(k,2k,2,2)$ & $k\times$16$^{d}$ & $2k\times$8$^{d}$ & a$\to$b\\
SSC$(2k,2k,3)$ & $2k\times$8$^{d}$ & $2k\times$8$^{d}$ & b\\
SC$(2k,4k,2,2)$ & $2k\times$8$^{d}$ & $4k\times$4$^{d}$ & b$\to$c\\
SSC$(4k,4k,3)$ & $4k\times$4$^{d}$ & $4k\times$4$^{d}$ & c\\
SC$(4k,16k,4,1)$ & $4k\times$4$^{d}$ & $16k\times$1$^{d}$ & c$\to$d\\
\bottomrule
\multicolumn{2}{c}{} & \multicolumn{2}{c}{}\\
\multicolumn{4}{c}{\sc NonConvNet Spatial Classifier}\\
\toprule
\bf Layer & \bf Input & \bf Output & \bf Sparsity\\
\midrule
DC$(16k,4k,4,1)$ & $16k\times1^{d}$ & $4k\times$4$^{d}$ & d$\to$c\\
DC$( 4k,2k,2,2)$ & $4k\times$4$^{d}$ & $2k\times$8$^{d}$ & c$\to$b\\
DC$( 2k, k,2,2)$ & $2k\times$8$^{d}$ & $k\times$16$^{d}$ & b$\to$a\\
\bottomrule
&&&\\
\end{tabular}
\end{minipage}
\begin{minipage}[t]{0.55\textwidth}
\begin{tabular}{lccl}
\multicolumn{4}{c}{\sc Decoder}\\
\toprule
\bf Layer & \bf Input & \bf Output & \bf Sparsity\\
\midrule
TC$(16k,4k,4,1)$ & $16k\times$1$^{d}$ & $4k\times$4$^{d}$ & d$\to$e\\
SSC$(4k,4k,3)$ & $4k\times$4$^{d}$ & $4k\times$4$^{d}$ & e\\
Sparsify & $4k\times$4$^{d}$ & $4k\times$4$^{d}$ & e$\to$f\\
SSC$(4k,4k,3)$ & $4k\times$4$^{d}$ & $4k\times$4$^{d}$ & f\\
TC$(4k,2k,2,2)$ & $4k\times$4$^{d}$ & $2k\times$8$^{d}$ & f$\to$g\\
SSC$(2k,2k,3)$ & $2k\times$8$^{d}$ & $2k\times$8$^{d}$ & g\\
Sparsify & $2k\times$8$^{d}$ & $2k\times$8$^{d}$ & g$\to$h\\
SSC$(2k,2k,3)$ & $2k\times$8$^{d}$ & $2k\times$8$^{d}$ & h\\
TC$(2k,k,2,2)$ & $2k\times$8$^{d}$ & $k\times$16$^{d}$ & h$\to$i\\
SSC$(k,k,3)$ & $k\times$16$^{d}$ & $k\times$16$^{d}$ & i\\
Sparsify & $k\times$16$^{d}$ & $k\times$16$^{d}$ & i$\to$j\\
SSC$(k,k,3)$ & $k\times$16$^{d}$ & $k\times$16$^{d}$ & j\\
\bottomrule
&&&\\
&&&\\
\end{tabular}
\end{minipage}
\mbox{
\hspace{20mm}
\includegraphics[width=0.11\paperwidth]{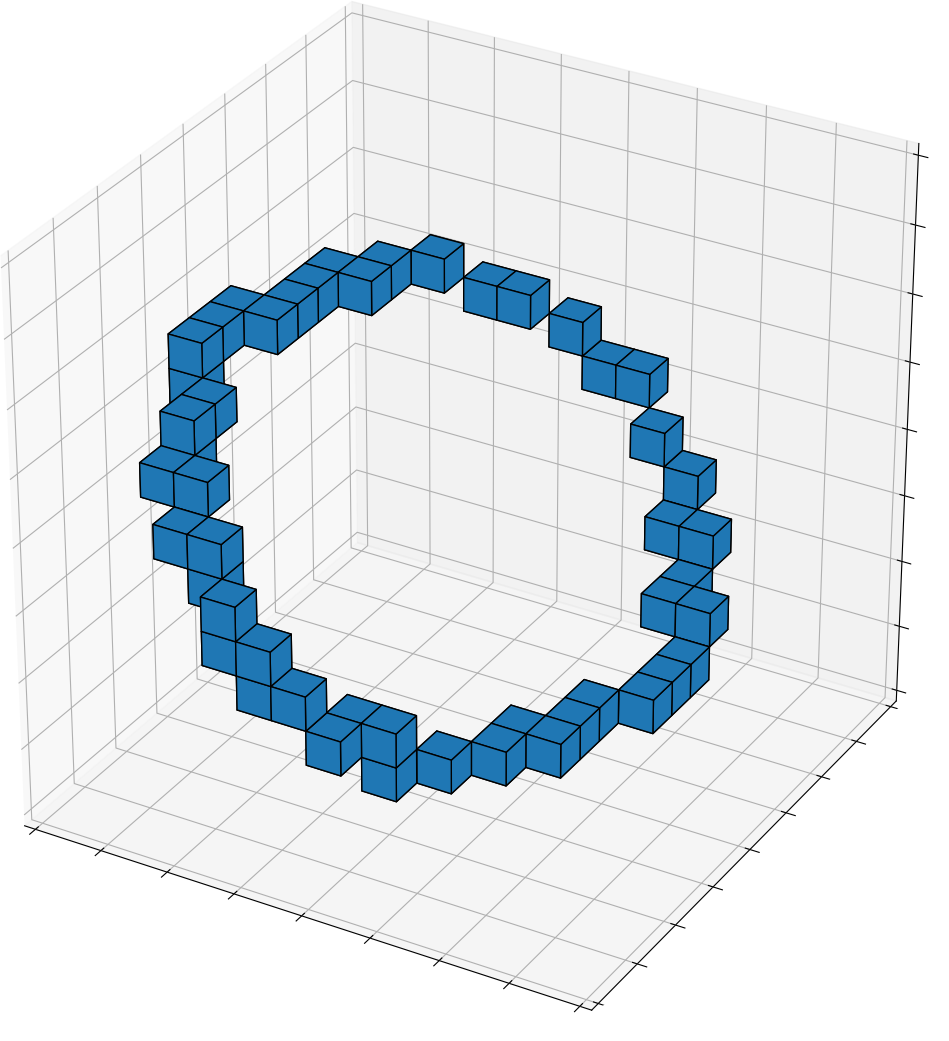}
\includegraphics[width=0.11\paperwidth]{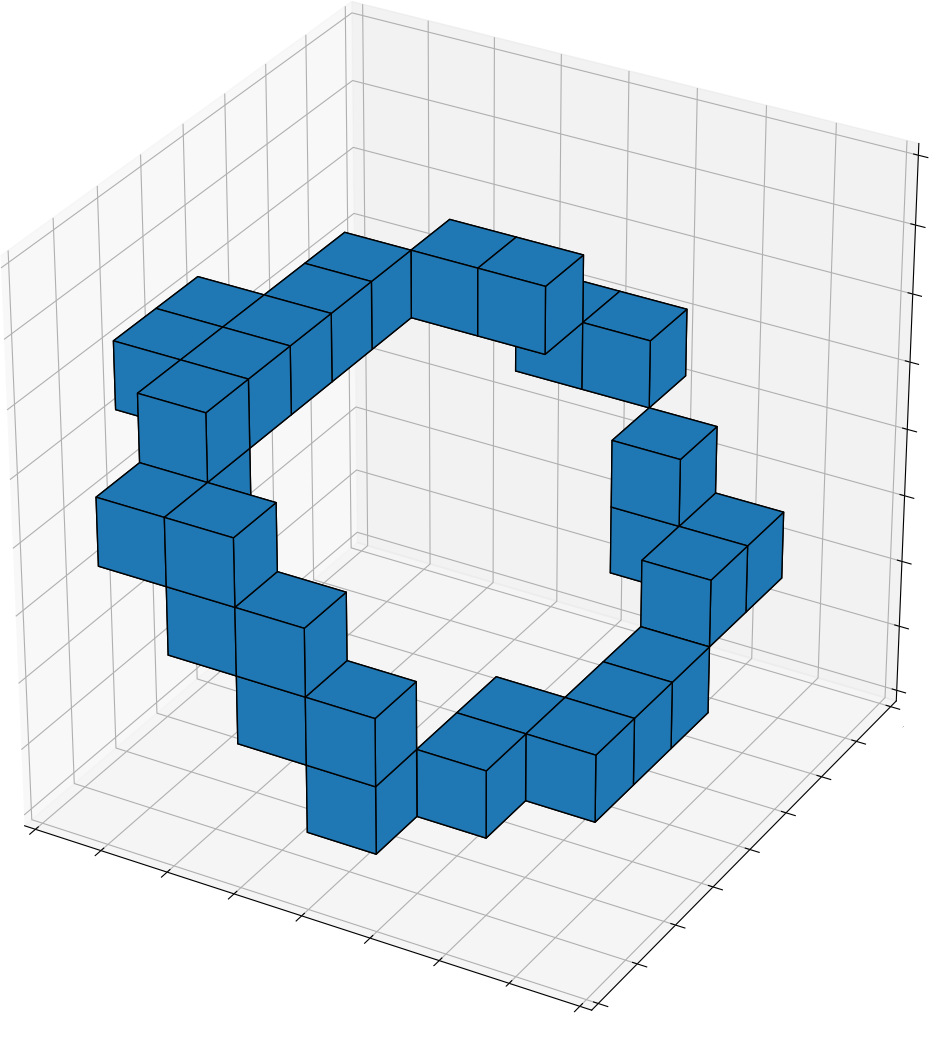}
\includegraphics[width=0.11\paperwidth]{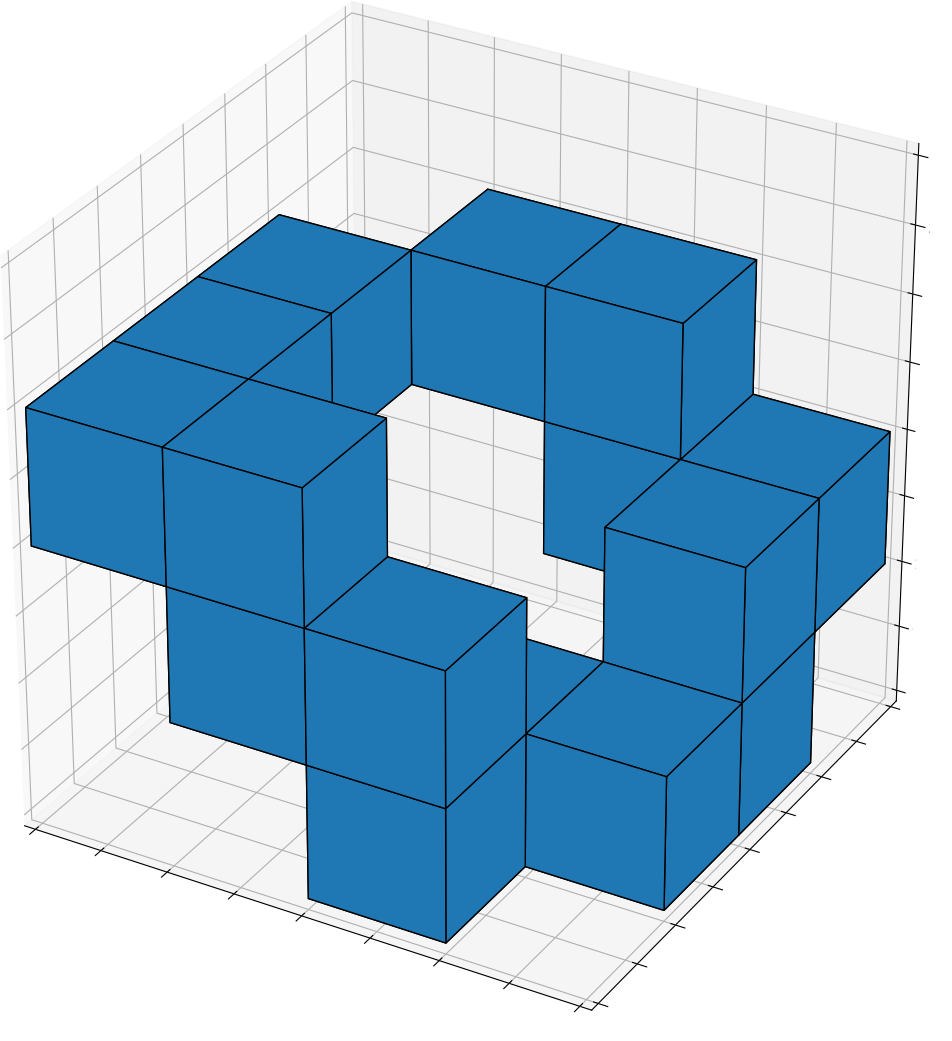}
\includegraphics[width=0.11\paperwidth]{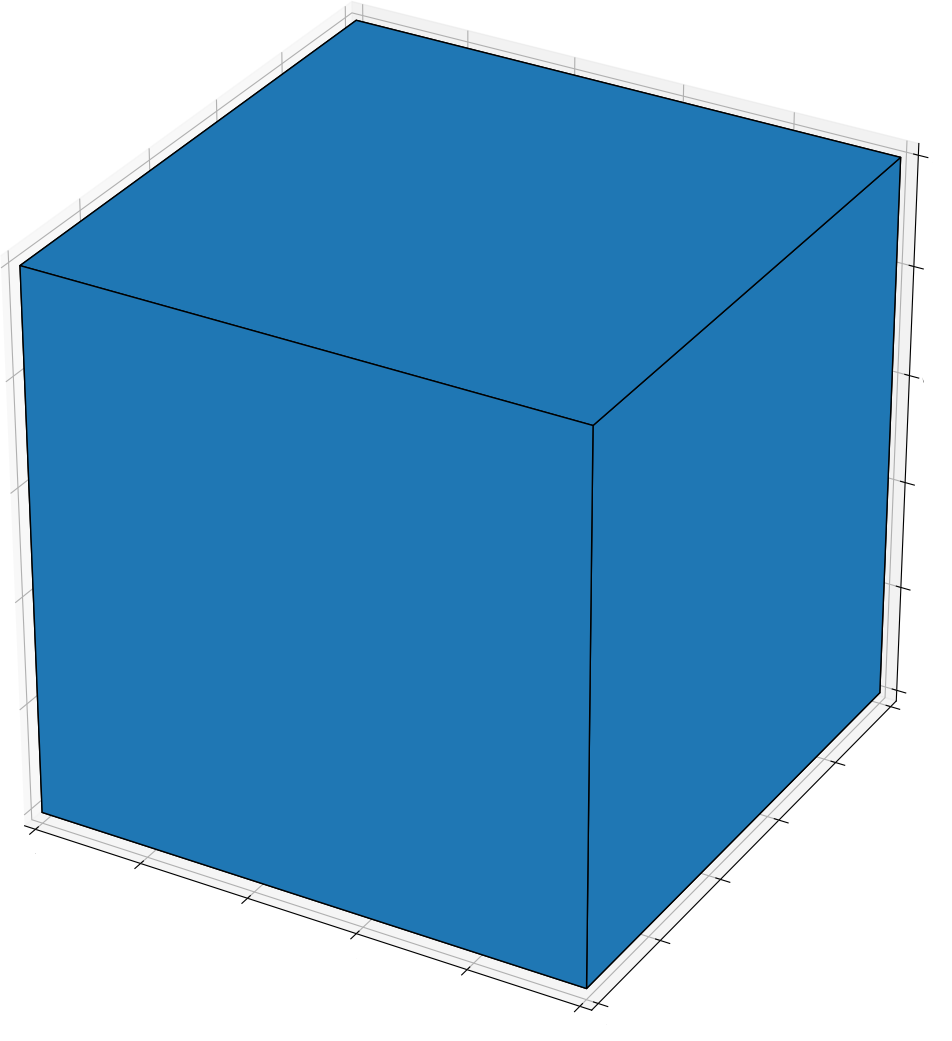}}\\
\mbox{\hspace{30mm}
{\bf
(a)\quad{}\quad{}\quad{}\quad{}\quad{}\quad{}
(b)\quad{}\quad{}\quad{}\quad{}\quad{}\quad{}
(c)\quad{}\quad{}\quad{}\quad{}\quad{}\quad{}
(d)}}\\
\includegraphics[width=0.11\paperwidth]{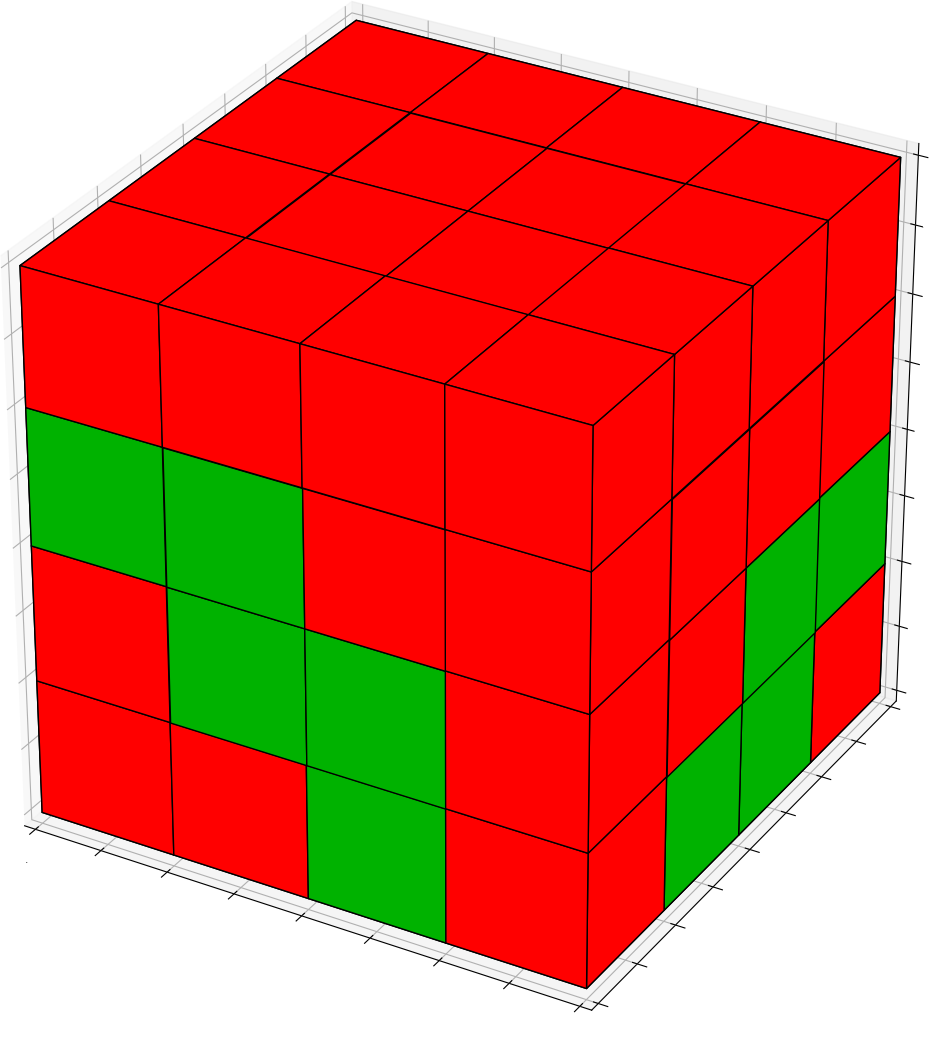}
\includegraphics[width=0.11\paperwidth]{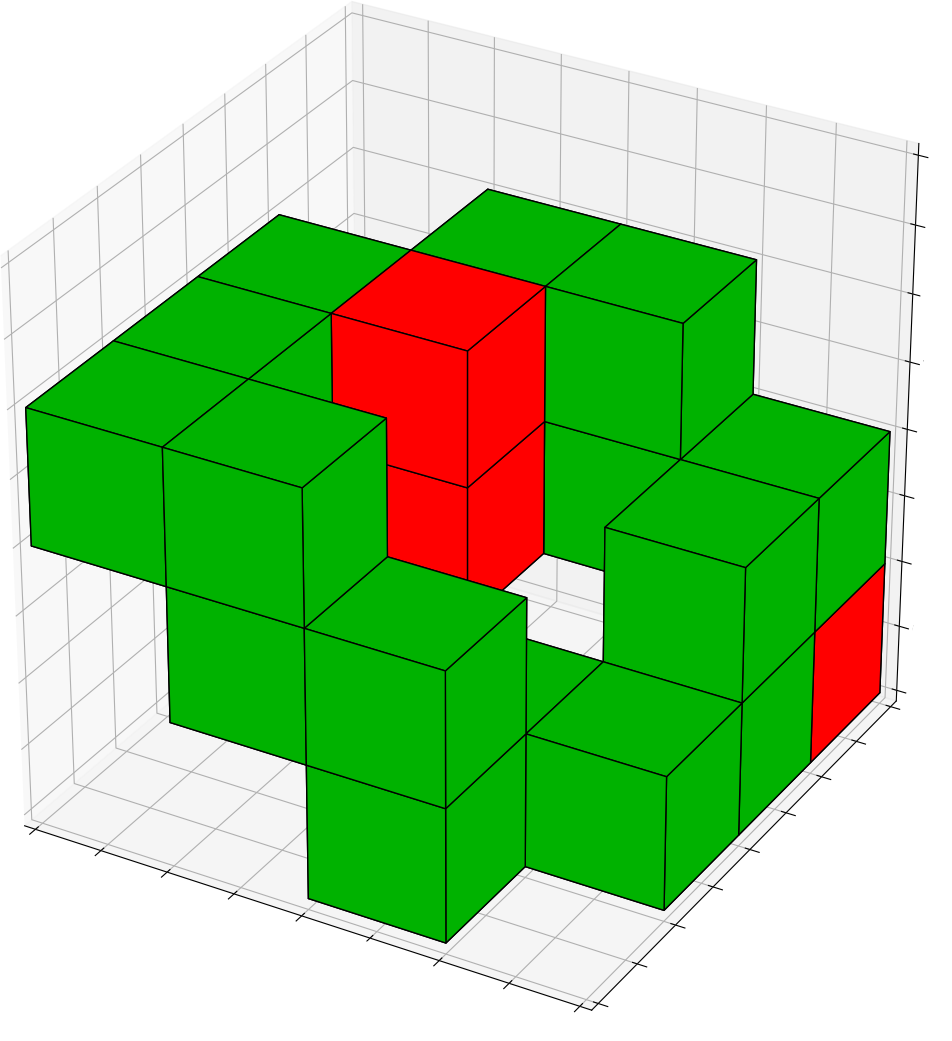}
\includegraphics[width=0.11\paperwidth]{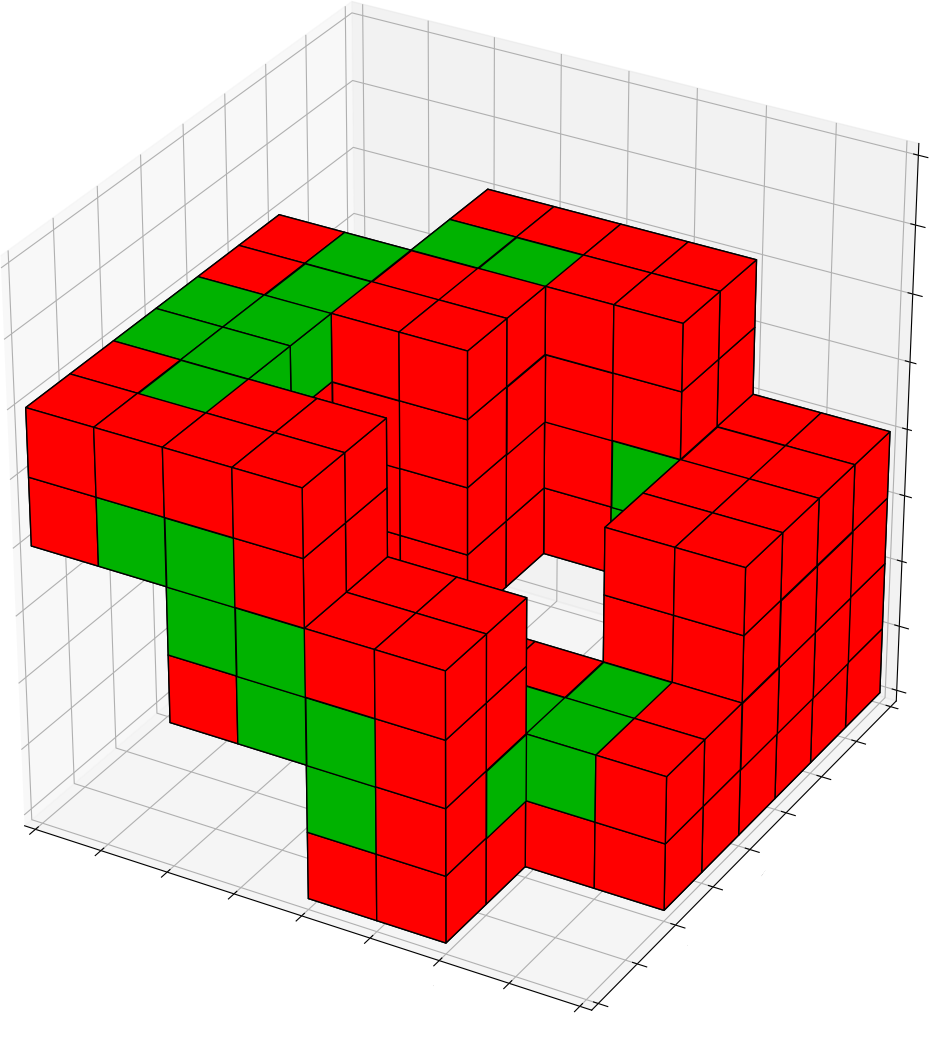}
\includegraphics[width=0.11\paperwidth]{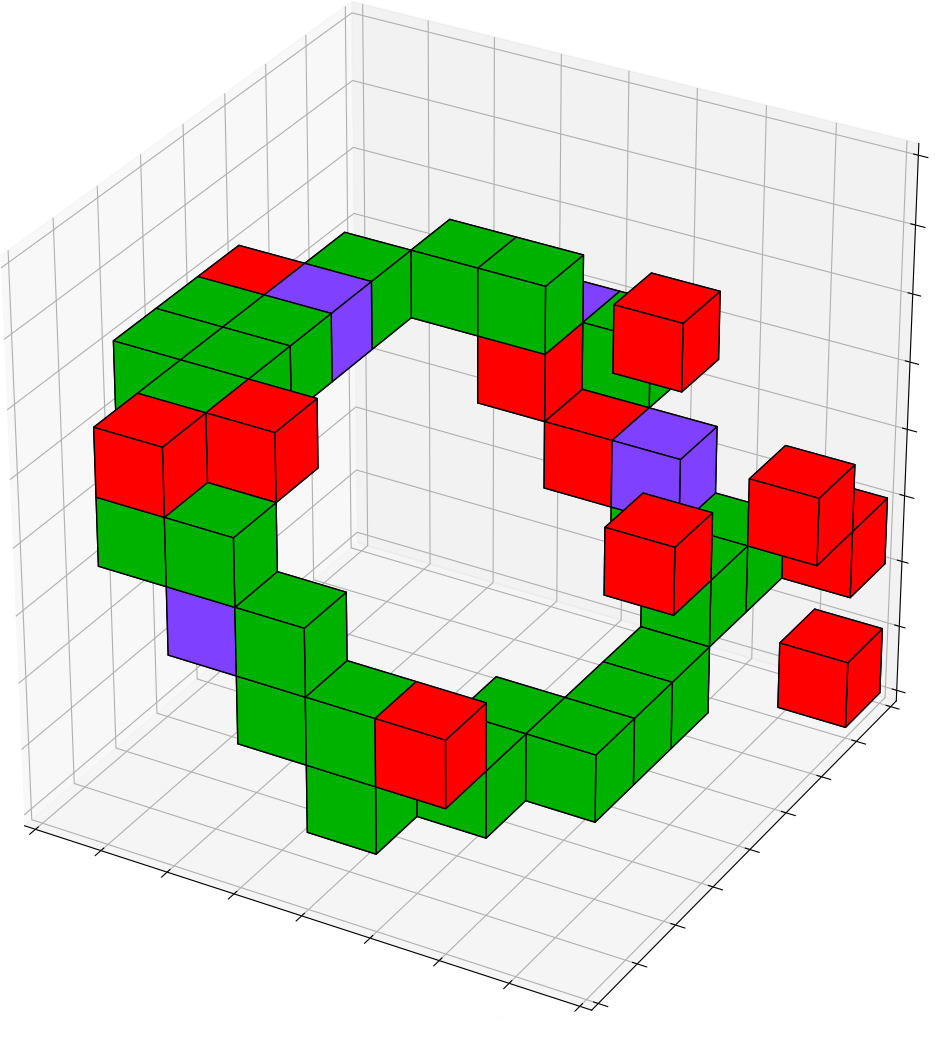}
\includegraphics[width=0.11\paperwidth]{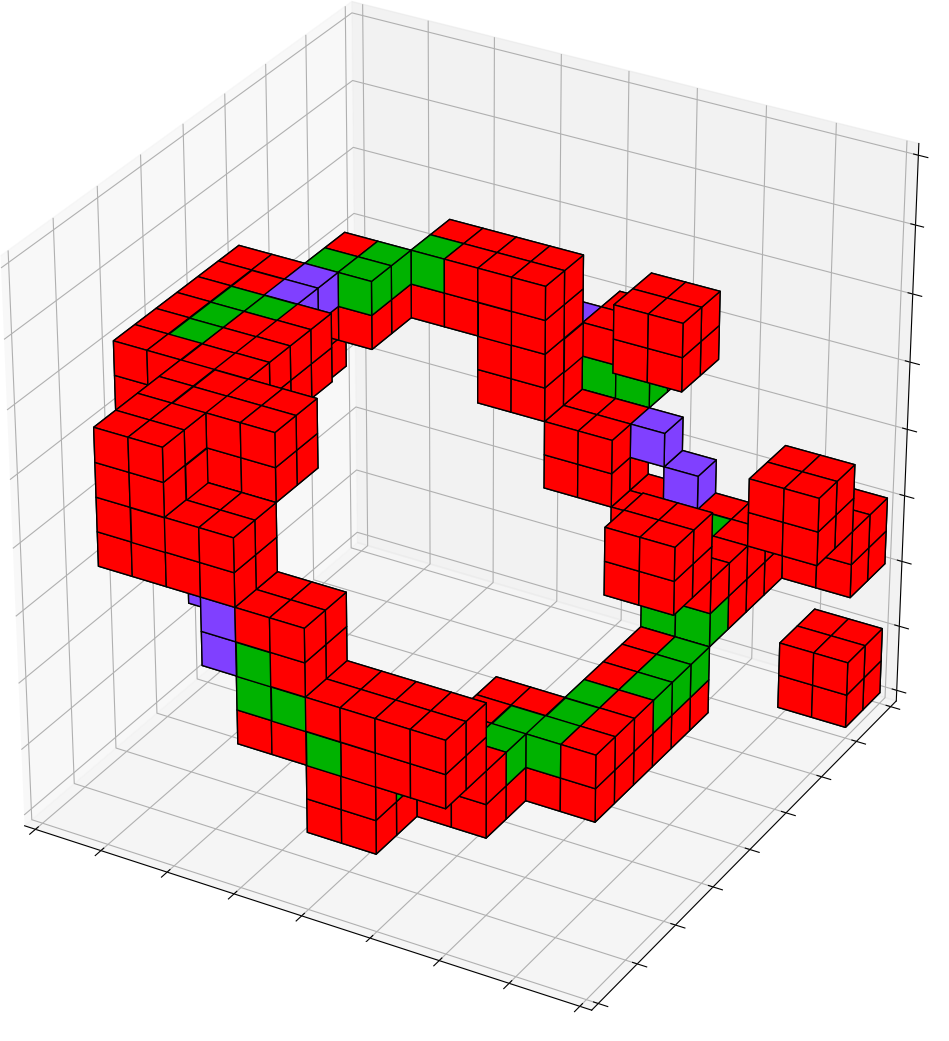}
\includegraphics[width=0.11\paperwidth]{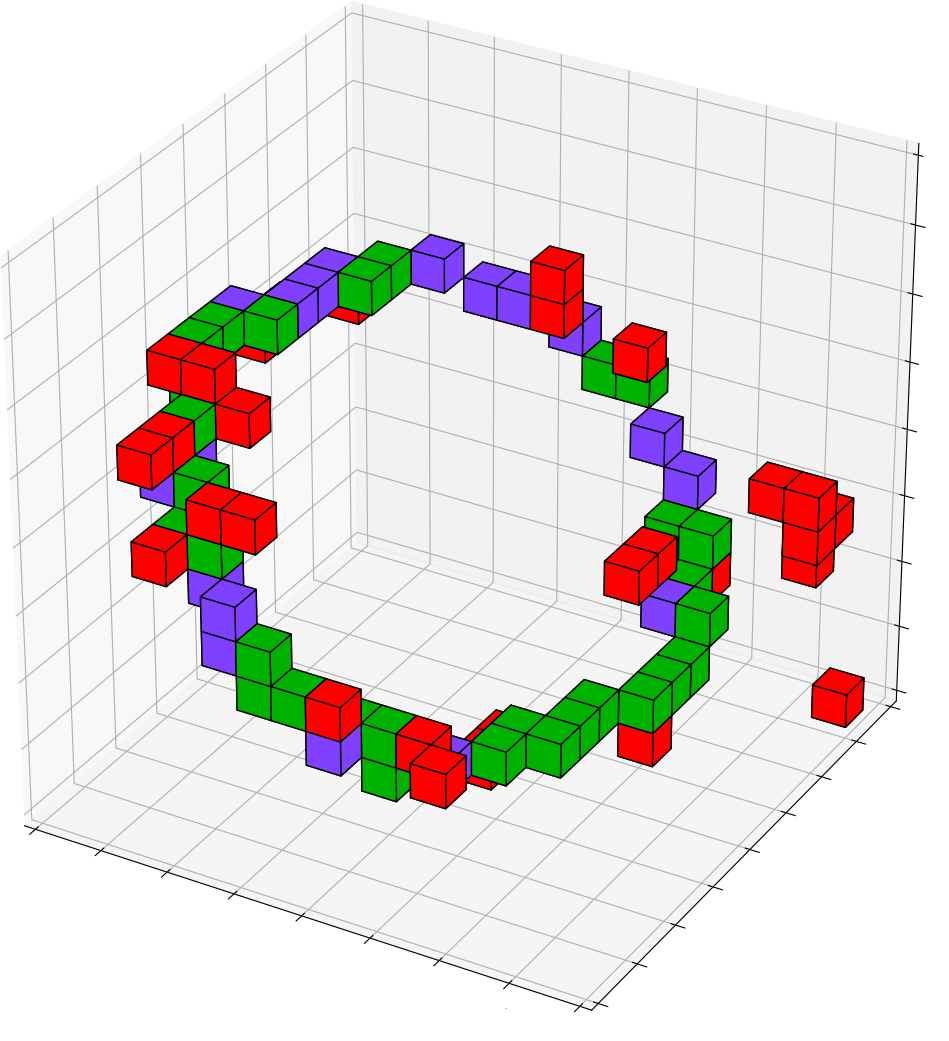}\\
\mbox{
\quad\quad{\bf
(e)
\quad{}\quad{}\quad{}\quad{}\quad{}\ \ \ 
(f)
\quad{}\quad{}\quad{}\quad{}\quad{}\ \ \ 
(g)
\quad{}\quad{}\quad{}\quad{}\quad{}\ \ \ 
(h)
\quad{}\quad{}\quad{}\quad{}\quad{}\ \ \ 
(i)
\quad{}\quad{}\quad{}\quad{}\quad{}\ \ \
(j)}}
\end{minipage}}
\end{center}
\caption{
\label{fig:arch}
{\bf Top:} Small sparse autoencoder architecture for inputs with spatial size $16^d$.
It can be expanded to process larger input, or modified to downsample the input by a fixed factor, i.e. $16\times$. {\bf Below:}
Illustration of the autoencoder operating on sparse input of size $16^3$.
Input {\bf (a)} is downscaled by strided convolutions to sizes {\bf (b)} $8^3$,
{\bf (c)} $4^3$ then {\bf (d)} $1^3$. During training, these patterns of active
sites form the ground truth for a hierarchical loss function.
At test time, reconstruction in the decoder is performed by alternating between `greedy' sparse transpose convolutions and sparsification layers; {\bf (e)} the scale increases to $4^3$  and {\bf (f)} some sites are deleted . This is repeated to take the scale to {\bf (g-h)} $8^3$ and finally  {\bf (i-j)} back
to $16^3$ . True positives are shown in green, false positives
in red, and false negatives in purple; true negatives are omitted.
To create the figures, sparsification decisions were made randomly
with 85\% accuracy. }
\end{figure*}

SSCN contain three types of layers:
\begin{description}
\item[SC$(m,n,f,s)$] Sparse convolution layers have $m$ input channels, $n$
output channels, filter size $f$ and stride $s$. They operate greedily:
if any site in the receptive field of an output square is active,
then the output is active. We set $f=s=2$ for downsampling by a factor
two.
\item[SSC$(m,n,f)$] Submanifold sparse convolutions always have stride
1. They preserve the sparsity structure, only being applied at sites
already active in the input. Deep SSCNs are primarily composed of
stacks of SSC convolutions, sometimes with skip connections added to produce simple ResNet style blocks \cite{ResNet}.
\item[DC$(m,n,f,s)$] Deconvolution layers restore the sparsity pattern
from a matching SC. They can be used to build U-Nets for image-to-image
mapping problems such as semantic segmentation.
\end{description}
To these, we add two new layers.
\begin{description}
\item[TC$(m,n,f,s)$] Transpose Convolutions will be used for upsampling.
Given the kernel size $f$, stride $s$, and input size $N^d$,
and the output size is $(fN-f+s)^d$. Upsampling is greedy, if an input location is active, all of the corresponding $f^d$ output locations are active.
\item[Sparsify] Sparsification layers convert some active spatial locations to in-active ones.
During training, they function like deconvolution layers, restoring the sparsity pattern to match the sparsity pattern at the same scale in the encoder.
During testing, if the first feature channel is positive, the site remains active and the feature channels pass unchanged from input to output. If the first feature channel is less than or equal to zero, the channel becomes non-active. In the special case of there being only one feature channel, this is equivalent to a ReLU activation function.
\end{description}
In the dense case, transpose convolution are
also known as fractional stride convolutions \cite{DCGAN} or `deconvolutions'.
With $r=s=2$, a TC operation corresponds to replacing each active site with a cube of size $2^d$.
Technically this preserves the level of sparsity. However, this is misleading; the volume has grown by a factor of $2^d$, but sparse sets of points typically have a fractal dimension of less than $d$, so we should expect greater sparsity. Looking at subfigures (c), (b) and (a) in Figure~\ref{fig:arch}, we see that the set of active sites corresponding to a 1-d circle in 3D space should become much more sparse as the scale increases. Hence the need for `sparsify' layers.

The framework is similar to Generative OctNets \cite{OctreeGeneratingNetwork, AdaptiveOCNN}, especially if TC and sparsifier layers are used back-to-back. However, separating the layers allows us to place trainable layers between upscaling and sparsifying. In Figure~\ref{fig:arch} we show an autoencoder operating on input size $16^3$.

\subsection{Encoder}
The encoder alternates between blocks of one or more SSC layers, and
downsampling SC layers. Each SC layer reduces the spatial size by a factor
of 2. Extra layers can be added to handle larger input. Once the spatial size is $4^d$, a
final SC layer can be used to reduce the spatial size
to a trivial $1^d$.

The sparsity patterns at different layers of the encoder are entirely
determined by the sparsity pattern in the input; they are independent of the encoder's trainable parameters.

\subsection{Decoder}
The decoder uses sequences of (i) a TC layer to upsample, (ii) an SSC layer to propagate information
spatially, (iii) a sparsify layer to increase sparsity, and (iv) an SSC layer to propagate information again before the next TC layer.

The spatial scales in the decoder, 1---4---8---16, are the inverse of the scales in the encoder.
During training, the sparsity pattern in the decoder after each sparsify layer is taken from the corresponding level of the encoder.
During testing, the sparsify layer keeps input locations where the first feature channel is positive, and deletes the rest.

\subsection{Hierarchical training loss}
To train the autoencoder, we define a loss that looks at the output features (unless the output is monochrome), and also each Sparsify layer.
During training, the output sparsity pattern matches the input sparsity pattern. The first term in our loss is the mean squared error of the reconstruction compared to the input over the set of input/output active spatial locations.
\begin{equation*}
\text{MSE}=\frac{1}{\#\text{active}}\sum_{x\text{ active}} \|\text{input}(x)-\text{output}(x)\|_2^2.
\end{equation*}
For each sparsifier layer, let $P$=`positive' denote the set of active sites in the corresponding layer of the encoder; let $N$=`negative' denote the set of inactive sites in the encoder.
Let $f=(f(x):x\in\mathbb{Z}^d)$
denote the first feature channel of the sparsifier layer input.
The sparsification loss for that sparsifier level is defined to be
\begin{equation*}
\sum_{x\in P}\max(1-f(x),0)^{2}+\sum_{x\in N}\max(1+f(x),0)^{2}.
\end{equation*}
This loss encourages the decoder to learn to
iteratively reproduce the sparsity pattern from the input. False positives, where a site is absent in the encoder but active in the decoder, can be corrected in later sparsification layers. However, false negatives, sites incorrectly turned off during decoding, cannot be corrected.

\subsection{Classifiers and NonConvNet spatial classifiers}
To make use of the latent space representations learnt by the sparse autoencoder, we need to be able to use the output of the encoder---the latent space---as input for downstream tasks such as classification and segmentation.

For classification, we can either have a linear layer followed by the softmax function. However, as the set of interesting classes may not be linearly separable in the latent space, we will also try training multilayer perceptrons (MLPs) as classifier; they will be fully connected neural networks with two hidden layers.

For segmentation, for each active point in the input, we want to produce a segmentation decision. However, for the results of the experiments to be meaningful, the classifier must not be allowed to base its decision on the input sparsity pattern, or else it could just ignore the latent space entirely and learn from the input from scratch.

To prevent this kind of cheating, we consider a `non-convolutional' decoder network, see the NonConvNet table in Figure~\ref{fig:arch}. It is implemented as a sparse ConvNet, but only using a sequence of $f=s=2$ deconvolutions. There is no overlap of the receptive fields, so given the latent vector, the segmentation decision at any input location is independent of the set of active input sites.

Compared to more typical decoder networks, the NonConvNet has some advantages. It is a small shallow network so it is quick to compute. It is easy to calculate the output at a particular location, without calculating the full output, e.g. `Is there a wall here?'
It is memory efficient in the sense that you can calculate the output without storing the autoencoder's input.
However, compared to other segmentation networks, such as U-Nets (see Section~\ref{subsec:U-Nets}), the lack of shortcut connections between the input and output will tend to limit accuracy when performing fine-grained segmentation; depending on the application this may be considered an acceptable trade-off.

\subsection{Arbitrary sized inputs}
The autoencoder in Figure~\ref{fig:arch} is designed to take input of a given size, $16^d$, and reduce it to a dimensionless latent vector with trivial spatial size $1^d$. The network can be expanded to take larger inputs, e.g. $64^d$, by adding extra $f=s=2$ SC/TC layers to the encoder/decoder, respectively. However, for large inputs such as scans of whole buildings, it is unrealistic to expect a single latent vector to capture all the information needed to reconstruct the extended scene.

A fixed size autoencoder could be applied to patches of the scene, to create a spatial ensemble of latent vectors. An advantage of this approach is that it is easy to update your `memory' when you revisit a location and find that the environment has changed.

Alternatively, and this is the approach taken here, one can build autoencoders that take arbitrary sized input $N^d$ and downsample by a fixed factor, i.e. by $16\times$, by adding extra $f=s=2$ SC/TC convolutions to the encoder/decoder, and removing the $f=4$, $s=1$ SC/TC convolutions. The latent space then has spatial size $(N/16)^d$.

When the latent space has a non-trivial spatial size, we will allow the segmentation classifier to consist of (a) an SSCN network operating on the latent space, followed by (b) a NonConvNet classifier. Storing just the latent space, or the output of (a), it is possible to evaluate the classifier at any input location.

\section{Experiments}
Our first experiments are with 2D handwriting datasets. In 2D, sparsity
is less important than in 3D or 4D, as the sparsity ratio will generally
be lower. However, it is interesting to look at datasets that are
relatively large compared to typical 3D/4D datasets, and to see if
the autoencoders can capture fine detail. We then look at two 3D segmentation dataset, and a 4D segmentation problem.

\subsection{Baselines\label{subsec:U-Nets}}

To assess the utility of the latent representations for other tasks, we will consider supervised and unsupervised baselines. We will pick networks
with similar computational cost to the encoder+classifier pairs. Methods trained fully supervised are marked with a $\bigstar$.

\begin{figure}
\begin{centering}
\includegraphics[width=0.8\columnwidth]{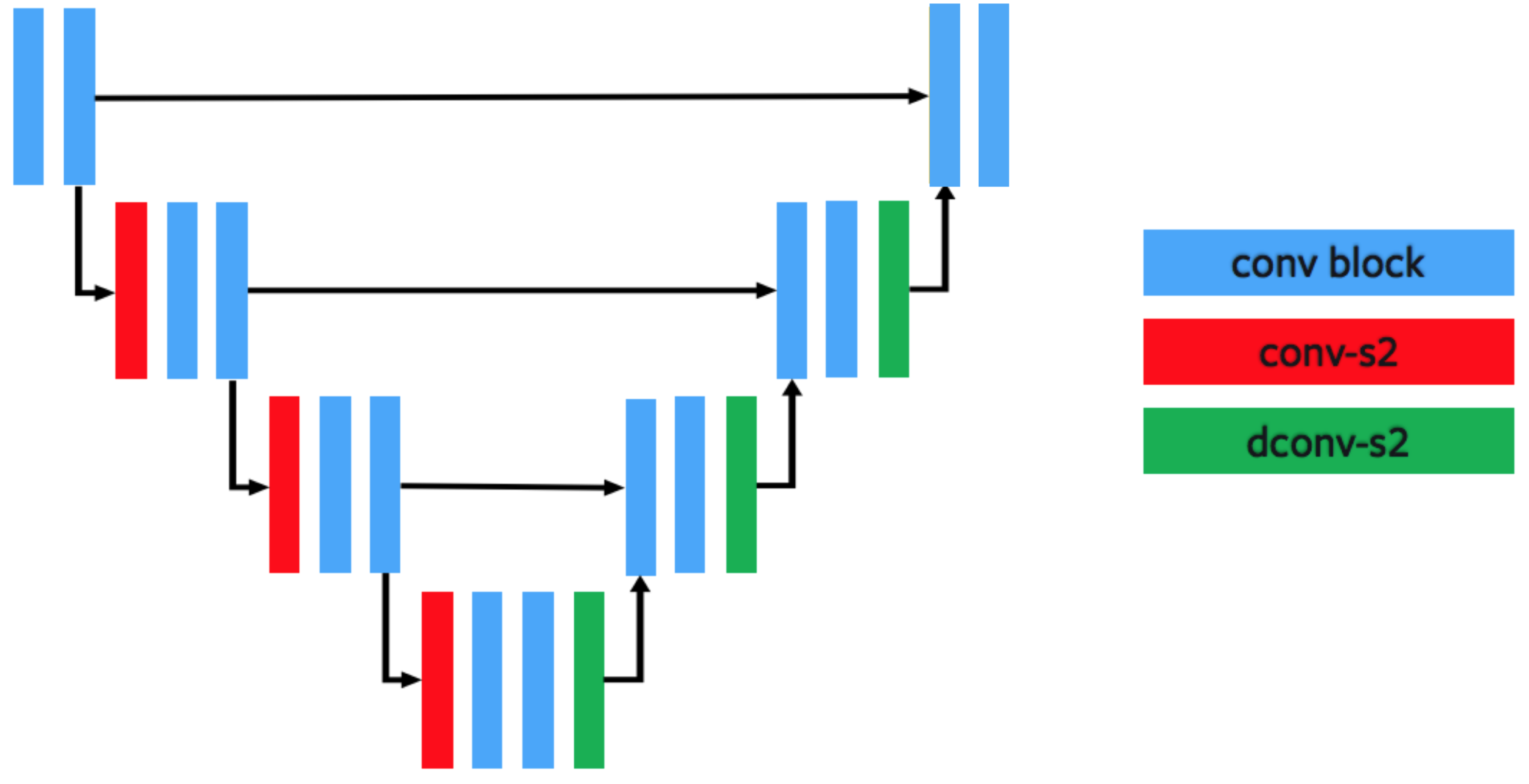}
\par\end{centering}
\caption{U-Net architecture for fully supervised segmentation. Blue blocks correspond to sparsity preserving SSC operations. The red blocks are stride-2 SC operations, and the green blocks are deconvolutions.\label{fig:unet}}
\end{figure}

\begin{description}
\item [Untrained]
As a simple baseline, we take a randomly initialized copy of the encoder \cite{conf/icml/SaxeKCBSN11}. To burn-in the batch norm population statistics, we perform 100 forward passes on the training data, but no actual training.
\item[Trained $\bigstar$] Another copy of the encoder, but trained fully supervised for the test task.
\item[U-Nets $\bigstar$] U-Nets have been applied to dense \cite{ronneberger2015unet} and sparse data \cite{graham2017submanifoldB} to obtain state-of-the-art results for segmentation problems. As they are trained fully supervised, with shortcut connections allowing segmentation decisions to be made with access to fine grained input detail, these provided an effective upper bound on the accuracy of unsupervised learning methods trained on the same number of examples. See Figure~\ref{fig:unet}.
\item[Shape Context]
Shape context features \cite{belongie2002shape} provide a simple summary of the local environment by performing pooling over a variety of scales. Let $n$ denote the number of input feature channels.
In parallel, the input is downscaled by average pooling by factors
of $1,2,4,\dots,2^{\ell-1}$. At each scale, at each active location, gather the $3^d$ feature vectors from neighboring spatial locations and concatenate them to produce $3^dn$ feature channels.
Unpooling the results from the different scales and concatenating them produces $3^dn\ell$ features at each active spatial location in the input.\\
For segmentation problems, the representation at each input-level spatial location
is fed into a multilayer perceptron (MLP) with 2 hidden layers to predict the voxel class.
\end{description}

\subsection{Handwriting in 2D space }
\begin{figure}
\centering{}\includegraphics[width=0.45\columnwidth]{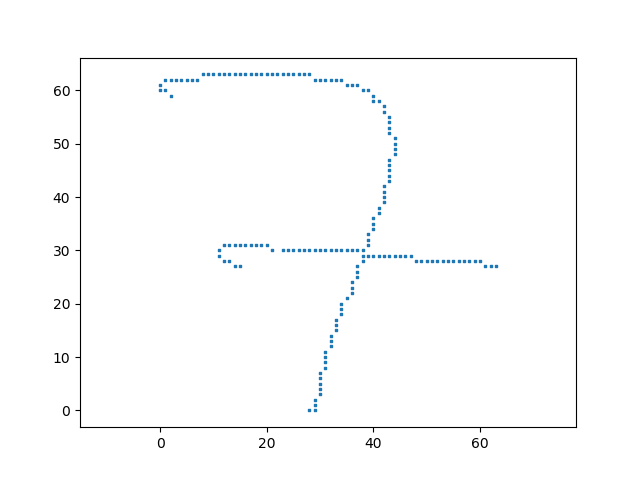}
\includegraphics[width=0.45\columnwidth]{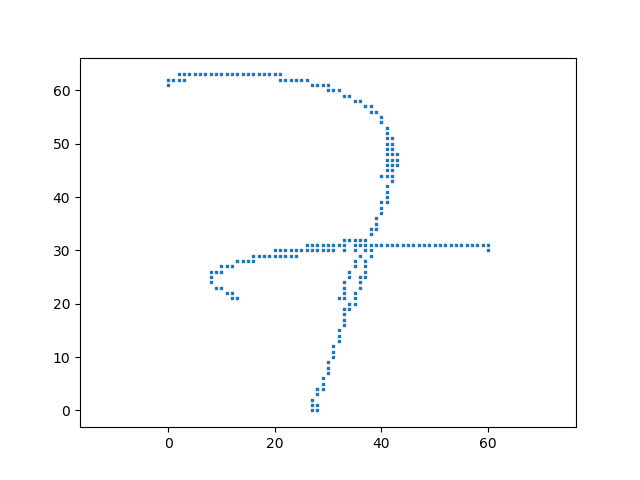}
\caption{Handwritten digit (left) and reconstruction (right). The reconstruction
seems to differ from the original by an elastic distortion. It is
far from the original in pixel space but still quite readable.\label{fig:7}}
\end{figure}

The PenDigits and Assamese handwriting datasets\footnote{\href{https://archive.ics.uci.edu/ml/datasets/}{https://archive.ics.uci.edu/ml/datasets}} contain samples of handwritten characters, each stored as a collection of paths in 2D space. The PenDigits dataset has samples of the digits $0,1,\dots,9$ with a total of 7494 training samples, and 3498 test samples, see Figure~\ref{fig:7}. The Assamese handwriting dataset has 45 samples of 183 characters; we split this into $36\times183$ training characters and $9\times183$ test characters.

We scale the input to size $64\times64$, and apply random affine transformation to the training data. For each dataset, we build 6 networks. Each network consists of an encoder network (c.f. Figure~\ref{fig:arch}) and on top of that either a linear classifier, or a 2-hidden-layer MLP. Each encoder is  either (i) randomly initialized, (ii) trained with full supervision, or (iii) trained unsupervised as part of a sparse autoencoder. The classifier is always trained with supervision. Results are in Table~\ref{tbl:hw}.

In the fully supervised case, the choice of classifier is unimportant; the encoder is already adapted to the character classes. The untrained encoder does significantly better than chance, especially with the MLP classifier. The encoder trained unsupervised as part of a sparse autoencoder does even better, performing only slightly worse than the fully supervised encoder on the PenDigits dataset.

\begin{table}
\begin{centering}
\begin{tabular}{llrr}
\toprule
\bf Dataset & \bf Encoder & \bf Linear & \bf MLP\\
\hline
\multirow{3}{*}{Digits} & Untrained & 16.84 & 6.26\\  
& Trained $\bigstar$& 1.14 & 0.89\\
& \bf Unsupervised$(64^d,k=16)$ & 2.80 & 1.26\\
\midrule
\multirow{3}{*}{Assamese} & Untrained & 68.43 & 44.51\\ 
& Trained $\bigstar$ & 2.79 & 2.61\\
& \bf Unsupervised$(64^d,k=32)$ & 28.05 & 16.51\\
\bottomrule
\end{tabular}
\par\end{centering}
\begin{centering}
\par\end{centering}
\caption{Handwriting recognition test errors, \%, for 10 and 183 class classification tasks. Within each column, the network architecture is the same, but trained differently. The classifier at the top of the network is either a linear layer or a 2-hidden layer fully-connected neural network.\label{tbl:hw}}
\end{table}

\subsection{ShapeNet 3D models}

\begin{figure}[t]
\begin{centering}
\includegraphics[width=1\columnwidth]{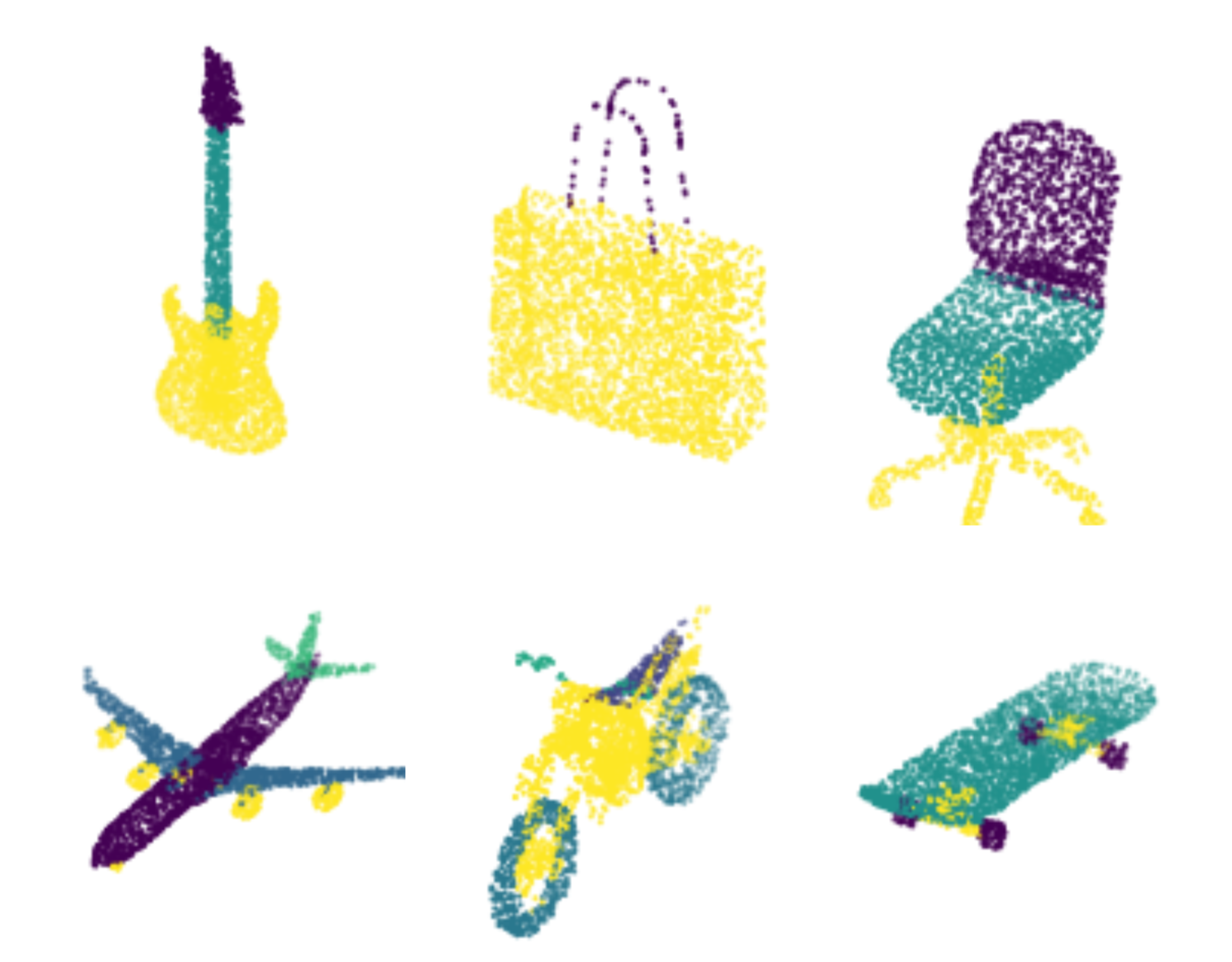}
\par\end{centering}
\caption{ShapeNet segmented point clouds. There are 16 object categories, each with 2-6 part types, e.g. a plane has wings, body, engines and a tail.\label{ShapeNetSegmented}}
\end{figure}

\begin{figure}[t]
\begin{centering}
\includegraphics[width=0.45\columnwidth]{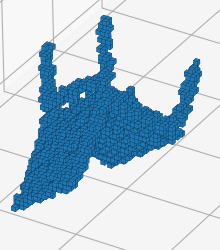}
\includegraphics[width=0.45\columnwidth]{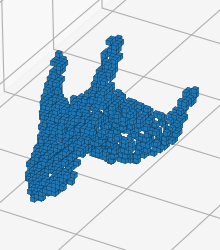}
\par\end{centering}
\caption{A randomly-oriented ShapeNet chair rendered with diameter 50 (left), and the reconstruction from an autoencoder with $16\times$ downscaling (right). The chair's style seems to have changed but location and pose are captured correctly.\label{ShapeNetRotated}}
\end{figure}

ShapeNet\footnote{\href{https://shapenet.cs.stanford.edu/iccv17/}{https://shapenet.cs.stanford.edu/iccv17/}}
is a dataset for semantic segmentation. There are 16 categories of object: airplane, bag, chair, etc. Each category is segmented into between 2 and 6 part types; see Figures~\ref{ShapeNetSegmented}.
Across all object categories, the dataset contains a total of 50 different object part classes. Each object is represented as a 3D point cloud obtained by sampling points uniformly from the surface of an underlying CAD model. Each point cloud contains between $2,000$ and $3,000$ points. We split the labeled data to obtain a training set with 6,955 examples and a validation set with 7,052 examples.

To make the reconstruction and segmentation problems more challenging, we randomly rotate the objects
in 3D; if airplanes always points along the $z$-axis, finding the
nose is rather easy, and you are limited to only ever fly in one direction!
Also, rather than treating the 16 object categories as separate tasks, we combine them.
We train the autoencoder on all categories. For the segmentation  task, we test  classification and segmentation ability simultaneously by treating the dataset as a 50 class segmentation problem (bag handle, plane wing, ...), and report the average intersection-over-union score (IOU).
We rendered the shapes at two different scales: diameter 15 in a grid of size $16^d$ and diameter 50 in an grid of size $64^d$.

At scale 15, we have a sparse autoencoder with input size $16^d$ to produce a latent representation with trivial size $1^d$. The baseline methods are shape context with an MLP of size 64, a U-Net, a randomly initialized encoder, and an encoder+NonConvNet pair trained end to end. See Table~\ref{tbl:3dSeg}.

For scale 50, we trained autoencoders that downscale space by $16\times$ and $32\times$. See Figure~\ref{ShapeNetRotated} and Table~\ref{tbl:3dSeg}.

\begin{table}
\begin{centering}
\begin{tabular}{clc}
\toprule
\bf Scale & \bf Method & \bf IOU\\
\midrule
\multirow{5}{*}{15}
&Shape Context & 0.134\\ 
&U-Net $\bigstar$        & 0.590\\  
&Untrained     & 0.161\\   
&Trained $\bigstar$   & 0.516\\  
&\bf Unsupervised$(16^d,k=32)$  & 0.278\\ 
\midrule
\multirow{4}{*}{50}
&Shape Context & 0.161\\ 
&U-Net $\bigstar$& 0.687\\ 
&\bf Unsupervised$(16\times,k=32)$ & 0.536 \\ 
&\bf Unsupervised$(32\times,k=32)$ & 0.420 \\ 
\bottomrule
\end{tabular}
\par\end{centering}
\begin{centering}
\par\end{centering}
\caption{ShapeNet segmentation results---average IOU over 50 classes. For scale 15, the latent space has trivial size $1^d$. For scale 50, it is downscaled by a factor of $16\times$ or $32\times$.\label{tbl:3dSeg}}
\end{table}

\begin{figure*}
\begin{centering}
\includegraphics[width=0.19\textwidth]{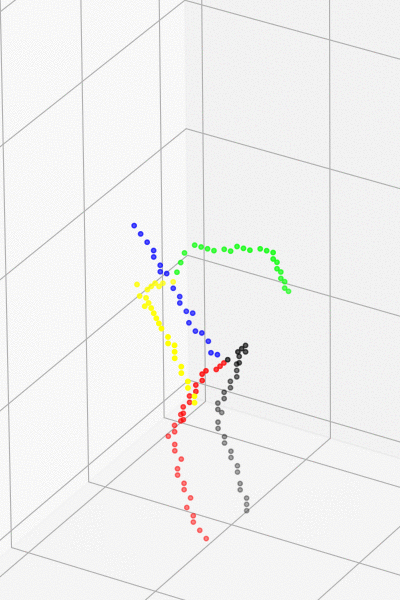}
\includegraphics[width=0.19\textwidth]{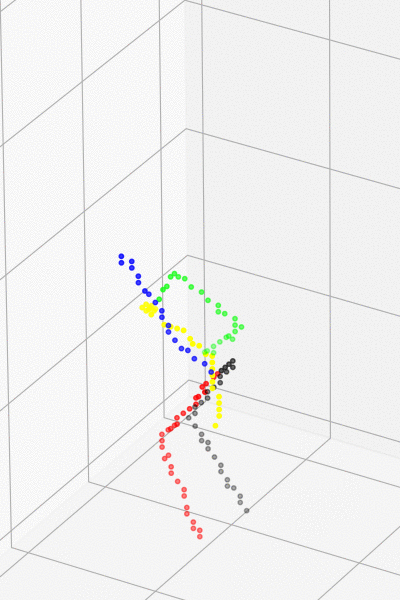}
\includegraphics[width=0.19\textwidth]{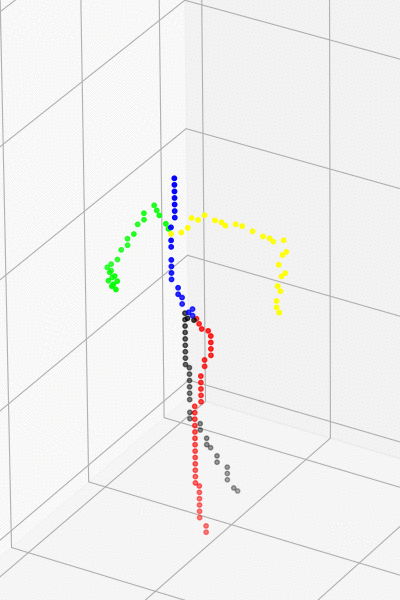}
\includegraphics[width=0.19\textwidth]{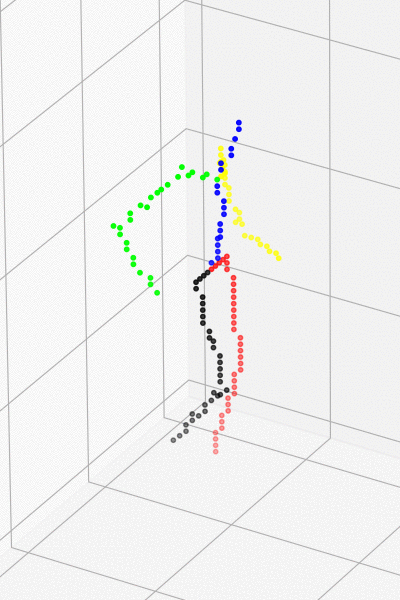}
\includegraphics[width=0.19\textwidth]{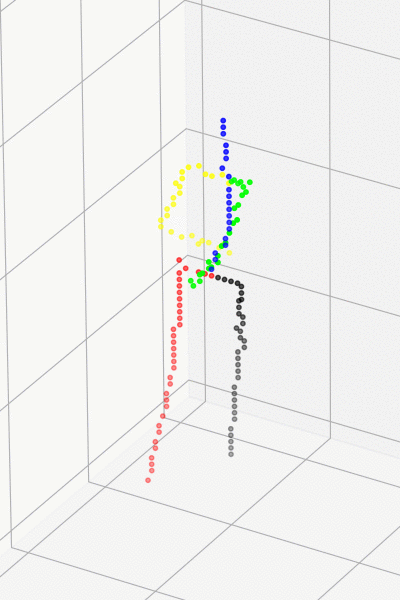}
\par\end{centering}
\caption{Skeleton wire frames from motion capture data: a person jumping and spinning. The 5 classes are left and right arms and legs, and the spine.\label{fig:mocap}}
\end{figure*}

\subsection{Motion capture walking wire frames}

The CMU Graphics Lab Motion Capture Database MOCAP\footnote{\href{http://mocap.cs.cmu.edu/}{http://mocap.cs.cmu.edu/}} contains keypoints data for people walking, running, dancing and doing gymnastics, see Figure~\ref{fig:mocap}.

We selected 1179 motion capture sequences for which we could extract complete and consistent set of keypoints, and used them to construct a simple wireframe model for the actors. The data can be represented as a simple time series, with the keypoint coordinates as features \cite{LearningMotionManifolds}, but this discards much of the 3D information. Instead we render the skeletons as 1+1 dimensional surfaces in 3+1 dimensional space-time (with one feature channel to indicate skeleton/not-skeleton). The model has no prior knowledge of how the skeleton is joined up or moves.

We split the dataset into 912 training sequences and 267 test sequences.
The method could in principle also be applied to motion capture data with multiple figures without modification to the sparse networks, but to simplify the data preparation, we restricted to the case of individual people.

We rendered samples of 64 frames (30 frames/s) in a cube of size $64^4$, and downscaled by a factor of $16\times$ or $32\times$. Baselines methods are 4D shape context features, a U-Net,  a randomly initialized network, and a fully supervised encoder+NonConvNet network.

For this experiment, we increased the number of features per enoder level linearly: e.g. 32, 64, 96, 128, 160, rather than by powers of 2. This is denoted `$k=32L$' in Table~\ref{tbl:4d}.

\begin{table}
\begin{centering}
\par\end{centering}
\begin{centering}
\begin{tabular}{lc}
\toprule
\bf Model & \bf IOU\\
\midrule
Shape Context & 0.718\\ 
U-Net$\bigstar$ & 0.988\\ 
Untrained & 0.701\\
Trained  $\bigstar$ & 0.913\\ 
\bf Unsupervised$(16\times,k=32L)$ & 0.879\\ 
\bf Unsupervised$(32\times,k=32L)$ & 0.808\\ 
\bottomrule
\end{tabular}
\par\end{centering}
\begin{centering}
\par\end{centering}
\caption{MOCAP 4D wireframe pose results with 5 classes.
\label{tbl:4d}}
\end{table}

\begin{figure*}[t]
\begin{centering}
\includegraphics[width=0.24\textwidth]{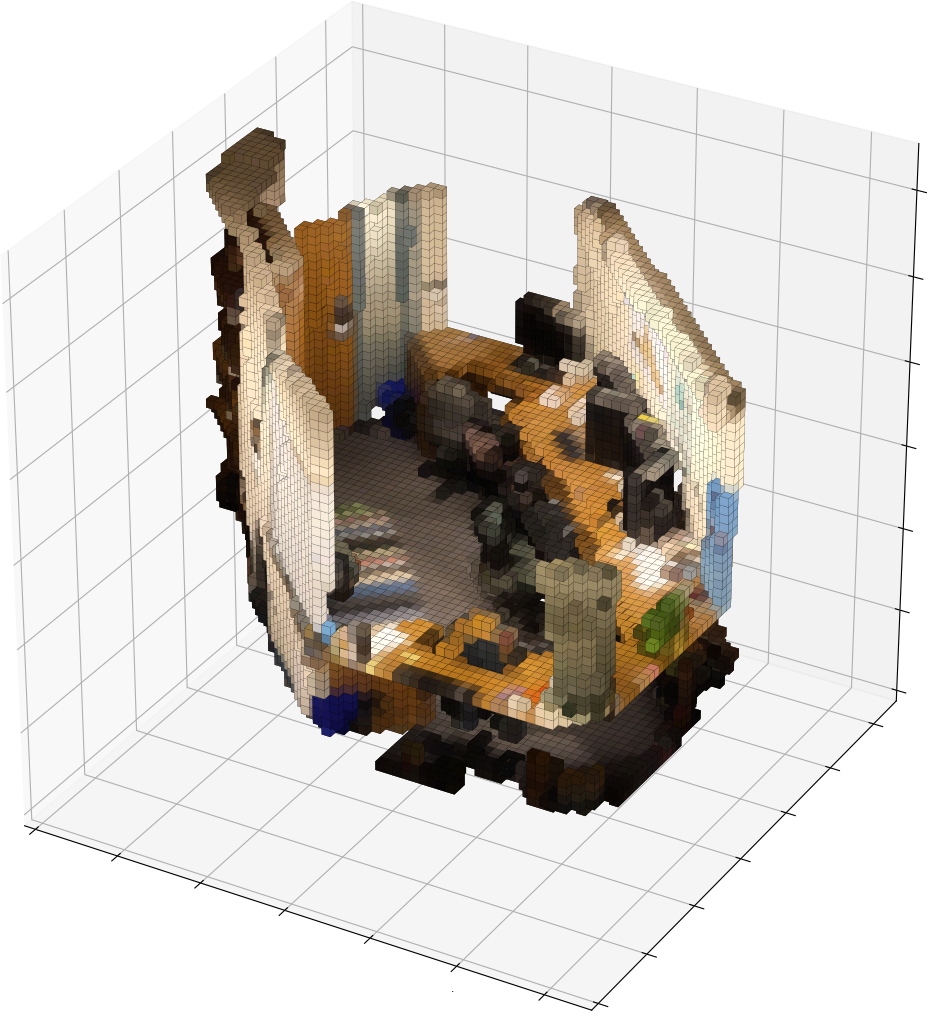}
\includegraphics[width=0.24\textwidth]{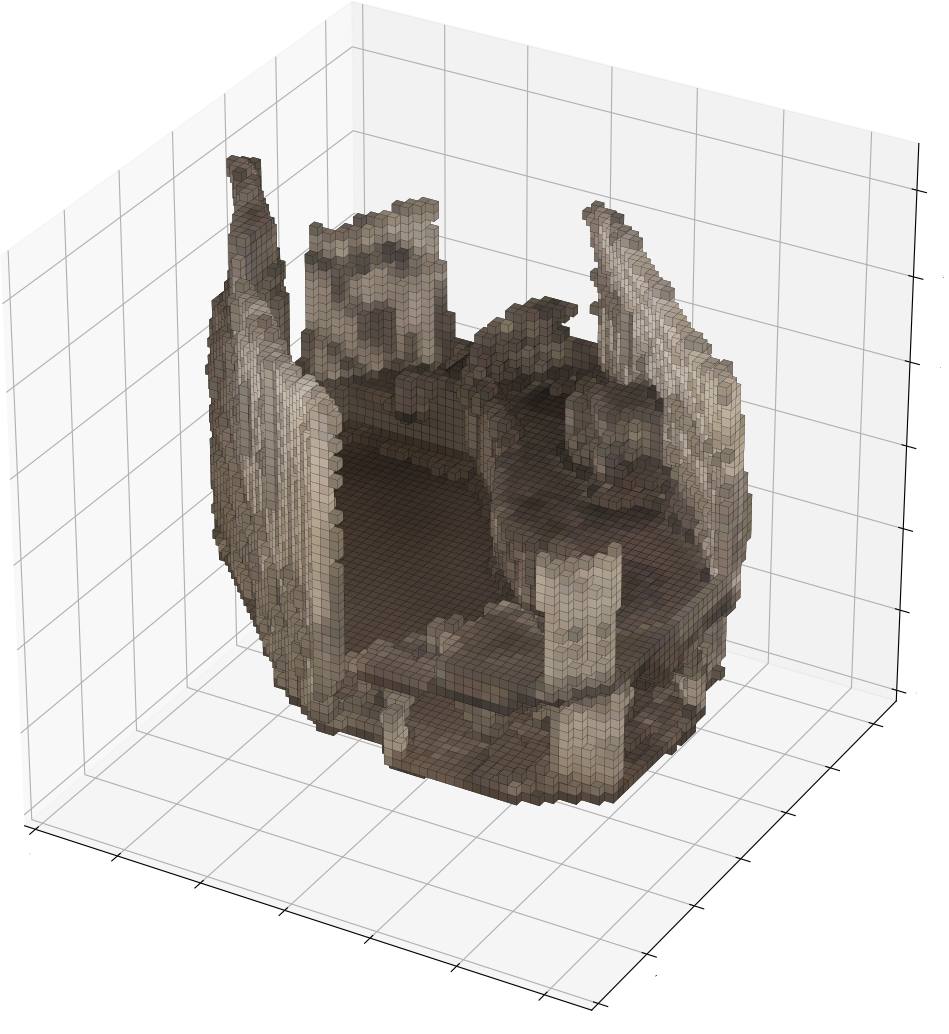}
\includegraphics[width=0.24\textwidth]{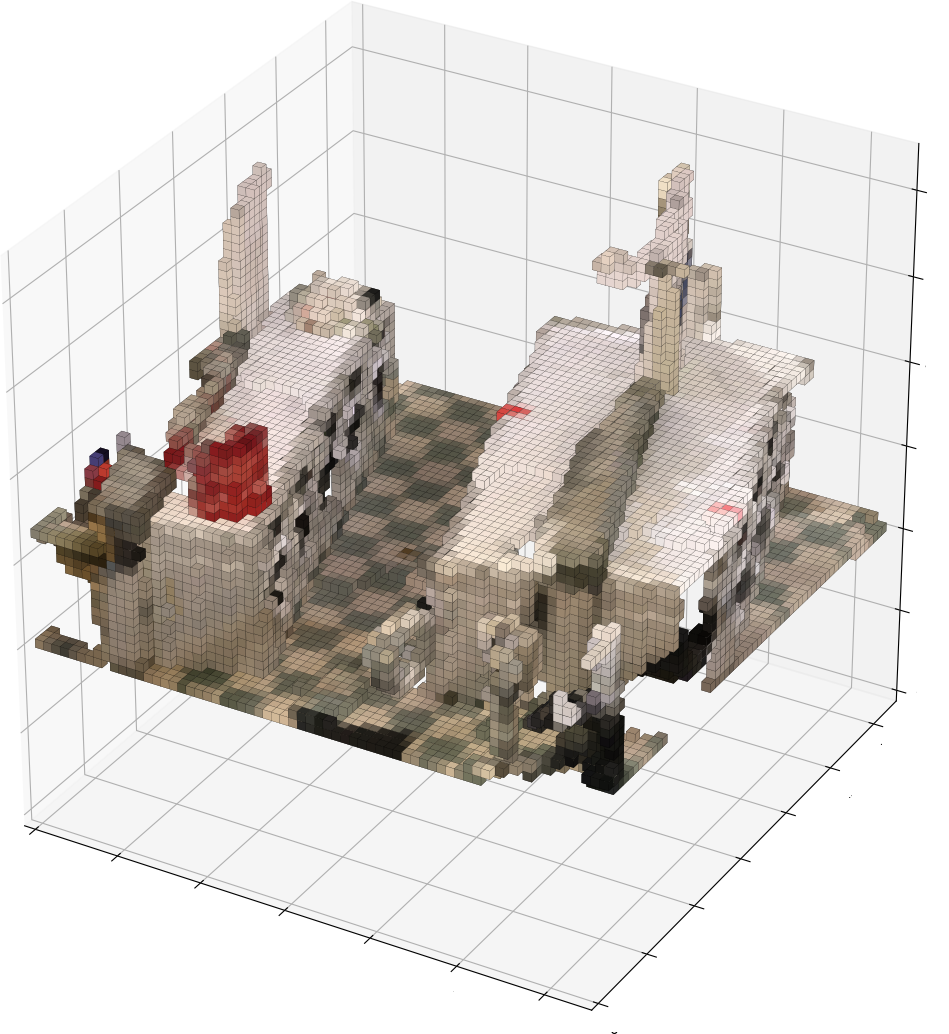}
\includegraphics[width=0.24\textwidth]{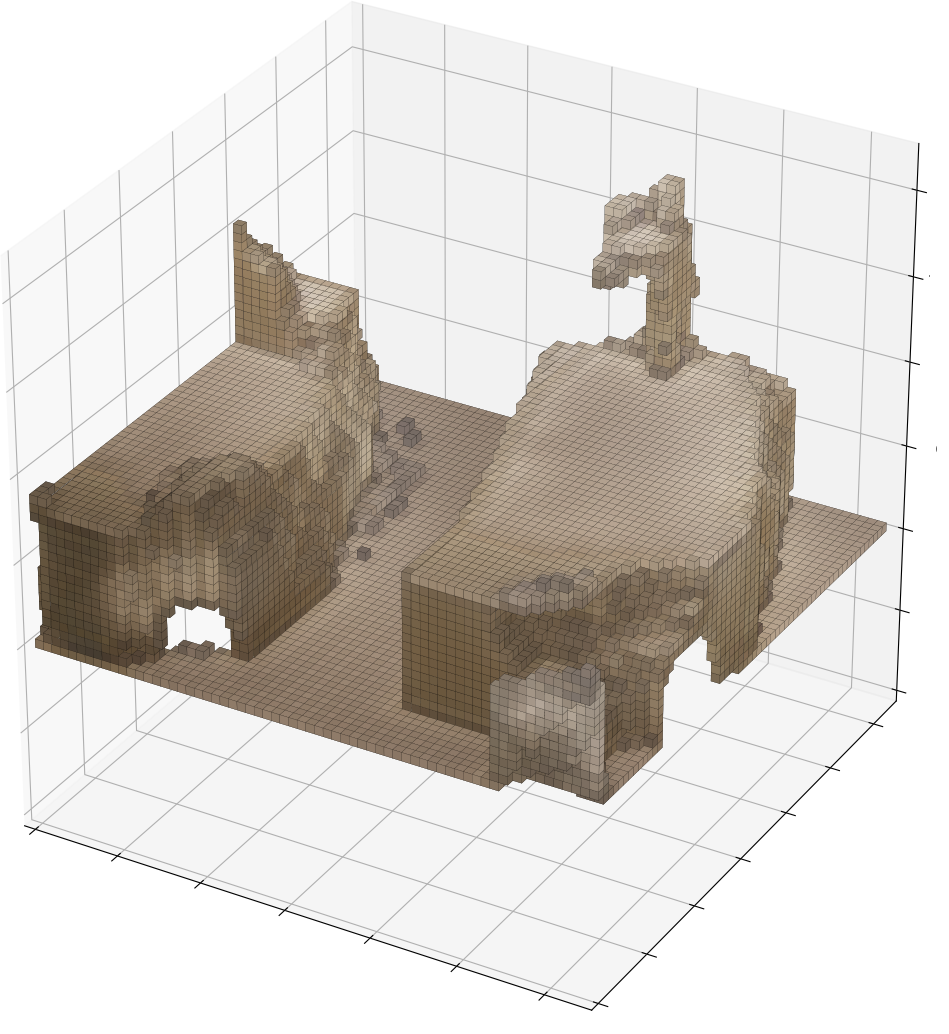}
\par\end{centering}
\caption{ScanNet RGB test scans and reconstructions from $16\times$ downsampled latent space. The reconstructions capture much of the shape, but little of the color information.\label{fig:scannet}}
\end{figure*}

\subsection{ScanNet room scenes}
The ScanNet dataset\footnote{\href{http://www.scan-net.org/}{http://www.scan-net.org/}}
has 1513 3D scans of scenes, segmented into 20 classes. We split the data into 1286 training samples and 227 test samples, see Figure~\ref{fig:scannet}.

For training we randomly rotate the scenes in the horizontal plane, and apply random affine data augmentation. We voxelize the the points with grid resolution $\sim7$cm, and use autoencoders to learn a latent space on a scale downsized by a factor of $8\times$, $16\times$ or $32\times$. For this experiment, we replaced the SSC blocks in Figure~\ref{fig:arch} with 2 simple ResNet block. This is denoted by `$k=32\times2R$' in Table~\ref{tbl:scannet}. Our results are calculated using 3-fold testing.

The fully supervised U-Net baseline is roughly on par with state-of-the-art methods \cite{ScanNetBenchmark}. The unsupervised encoder compares respectably to some of the fairly recent fully supervised methods.

We repeated the supervised learning using only 10\% of the labelled scenes, see Figure~\ref{tbl:scannetSmallData}. The gap between the fully supervised U-Net reduces. The unsupervised representation outperforms an equivalent network trained fully supervised on the reduced training set.

\begin{table}
\begin{centering}
\par\end{centering}
\begin{centering}
\begin{tabular}{ll}
\toprule
\bf Method & \bf IOU\\
\midrule
Shape Context & 0.211\\
U-Net$\bigstar$ & 0.703\\
3DMV $\bigstar$ \cite{3DMV} & 0.484$\dagger$\\
SurfaceConvPF$ \bigstar$\cite{journals/corr/abs-1808-04952}   & 0.442$\dagger$\\
Mink34 $\bigstar$\cite{ScanNetBenchmark} & 0.679$\dagger$ \\
\bf Unsupervised$(\,\,\,8\times,k=32\times2R)$  & 0.518\\
\bf Unsupervised$(16\times,k=32\times2R)$ & 0.414\\
\bf Unsupervised$(32\times,k=32\times2R)$ & 0.299\\
\bottomrule
\end{tabular}
\par\end{centering}
\caption{ScanNet room segmentation results.\\ $\dagger$$=$Cited results were calculated on a different test set. \label{tbl:scannet}}
\end{table}

\begin{table}
\begin{centering}
\par\end{centering}
\begin{centering}
\begin{tabular}{lc}
\toprule
\bf Method& \bf IOU\\
\midrule
Shape Context & 0.172\\
U-Net $\bigstar$& 0.460\\
Trained
$\bigstar$ & 0.212\\
\bf Unsupervised$(16\times,k=32\times2R)$ & 0.295\\
\bottomrule
\end{tabular}
\par\end{centering}
\caption{ScanNet using 10\% of the training labels.\label{tbl:scannetSmallData}}
\end{table}

\section{Conclusion}

We have introduced a new framework for building spatially-sparse autoencoder networks in 2D, 3D and 4D.
We have also introduced a number of segmentation benchmark tasks to measure the quality of the latent space representations generated by the autoencoders. Other possible uses include reinforcement learning tasks related to navigation in 3D environments \cite{House3D} and embodied Q\&A\footnote{\href{https://embodiedqa.org/}{https://embodiedqa.org/}}.


\begin{thebibliography}{10}

\bibitem{ScanNetBenchmark}
Scannet benchmark challenge.
\newblock
  \href{http://kaldir.vc.in.tum.de/scannet_benchmark/}{http://kaldir.vc.in.tum.de/scannet\_benchmark/}.
\newblock Accessed: 2018-11-16.

\bibitem{belongie2002shape}
Serge Belongie, Jitendra Malik, and Jan Puzicha.
\newblock {Shape Matching and Object Recognition using Shape Contexts}.
\newblock {\em IEEE Transactions on Pattern Analysis and Machine Intelligence},
  2002.

\bibitem{predictingNoise}
Piotr Bojanowski and Armand Joulin.
\newblock Unsupervised learning by predicting noise.
\newblock In Doina Precup and Yee~Whye Teh, editors, {\em Proceedings of the
  34th International Conference on Machine Learning, ICML 2017, Sydney, NSW,
  Australia, 6-11 August 2017}, volume~70 of {\em Proceedings of Machine
  Learning Research}, pages 517--526. PMLR, 2017.

\bibitem{journals/corr/abs-1807-05520}
Mathilde Caron, Piotr Bojanowski, Armand Joulin, and Matthijs Douze.
\newblock Deep clustering for unsupervised learning of visual features.
\newblock {\em CoRR}, abs/1807.05520, 2018.

\bibitem{3DMV}
Angela Dai and Matthias Nie{\ss}ner.
\newblock 3dmv: Joint 3d-multi-view prediction for 3d semantic scene
  segmentation.
\newblock {\em CoRR}, abs/1803.10409, 2018.

\bibitem{exemplar}
Alexey Dosovitskiy, Jost~Tobias Springenberg, Martin~A. Riedmiller, and Thomas
  Brox.
\newblock Discriminative unsupervised feature learning with convolutional
  neural networks.
\newblock {\em CoRR}, abs/1406.6909, 2014.

\bibitem{engelcke2017vote3deep}
Martin Engelcke, Dushyant Rao, Dominic~Zeng Wang, Chi~Hay Tong, and Ingmar
  Posner.
\newblock {Vote3Deep: Fast Object Detection in 3D Point Clouds using Efficient
  Convolutional Neural Networks}.
\newblock {\em IEEE International Conference on Robotics and Automation}, 2017.

\bibitem{graham2015sparse}
Benjamin Graham.
\newblock {Sparse 3D Convolutional Neural Networks}.
\newblock {\em British Machine Vision Conference}, 2015.

\bibitem{graham2017submanifoldB}
Benjamin Graham, Martin Engelcke, and Laurens van~der Maaten.
\newblock {3D Semantic Segmentation with Submanifold SparseConvNets}.
\newblock 2017.
\newblock http://arxiv.org/abs/1711.10275.

\bibitem{ResNet}
Kaiming He, Xiangyu Zhang, Shaoqing Ren, and Jian Sun.
\newblock {Identity Mappings in Deep Residual Networks}.
\newblock {\em European Conference on Computer Vision}, 2016.

\bibitem{LearningMotionManifolds}
Daniel Holden, Jun Saito, Taku Komura, and Thomas Joyce.
\newblock Learning motion manifolds with convolutional autoencoders.
\newblock In {\em SIGGRAPH Asia Technical Briefs}, pages 18:1--18:4. ACM, 2015.

\bibitem{FastTextClassification}
Armand Joulin, Edouard Grave, Piotr Bojanowski, and Tomas Mikolov.
\newblock Bag of tricks for efficient text classification.
\newblock In {\em Proceedings of the 15th Conference of the European Chapter of
  the Association for Computational Linguistics: Volume 2, Short Papers}, pages
  427--431. Association for Computational Linguistics, April 2017.

\bibitem{PermutohedralLatticeCNN}
Martin Kiefel, Varun Jampani, and Peter~V. Gehler.
\newblock Permutohedral lattice cnns.
\newblock {\em ICLR}, 2015.

\bibitem{klokov2017escape}
Roman Klokov and Victor Lempitsky.
\newblock {Escape from Cells: Deep Kd-Networks for The Recognition of 3D Point
  Cloud Models}.
\newblock {\em arXiv preprint arXiv:1704.01222}, 2017.

\bibitem{FaderNetworks}
Guillaume Lample, Neil Zeghidour, Nicolas Usunier, Antoine Bordes, Ludovic
  DENOYER, and Marc'Aurelio Ranzato.
\newblock Fader networks:manipulating images by sliding attributes.
\newblock In I.~Guyon, U.~V. Luxburg, S.~Bengio, H.~Wallach, R.~Fergus,
  S.~Vishwanathan, and R.~Garnett, editors, {\em Advances in Neural Information
  Processing Systems 30}, pages 5967--5976. Curran Associates, Inc., 2017.

\bibitem{lecun-98}
Y.~LeCun, L.~Bottou, Y.~Bengio, and P.~Haffner.
\newblock Gradient-based learning applied to document recognition.
\newblock {\em Proceedings of the IEEE}, 86(11):2278--2324, November 1998.

\bibitem{long2015fully}
Jonathan Long, Evan Shelhamer, and Trevor Darrell.
\newblock {Fully Convolutional Networks for Semantic Segmentation}.
\newblock {\em Proceedings of the IEEE Conference on Computer Vision and
  Pattern Recognition}, 2015.

\bibitem{Word2Vec}
Tomas Mikolov, Ilya Sutskever, Kai Chen, Greg~S Corrado, and Jeff Dean.
\newblock Distributed representations of words and phrases and their
  compositionality.
\newblock In C.~J.~C. Burges, L.~Bottou, M.~Welling, Z.~Ghahramani, and K.~Q.
  Weinberger, editors, {\em Advances in Neural Information Processing Systems
  26}, pages 3111--3119. Curran Associates, Inc., 2013.

\bibitem{jigsaw}
Mehdi Noroozi and Paolo Favaro.
\newblock Unsupervised learning of visual representations by solving jigsaw
  puzzles.
\newblock {\em CoRR}, abs/1603.09246, 2016.

\bibitem{journals/corr/abs-1808-04952}
Hao Pan, Shilin Liu, Yang~Liu 0014, and Xin~Tong 0001.
\newblock Convolutional neural networks on 3d surfaces using parallel frames.
\newblock {\em CoRR}, abs/1808.04952, 2018.

\bibitem{qi2016pointnet}
Charles~R Qi, Hao Su, Kaichun Mo, and Leonidas~J Guibas.
\newblock {PointNet: Deep Learning on Point Sets for 3D Classification and
  Segmentation}.
\newblock {\em arXiv preprint arXiv:1612.00593}, 2016.

\bibitem{DCGAN}
Alec Radford, Luke Metz, and Soumith Chintala.
\newblock Unsupervised representation learning with deep convolutional
  generative adversarial networks.
\newblock {\em CoRR}, abs/1511.06434, 2015.

\bibitem{riegler2016octnet}
Gernot Riegler, Ali~Osman Ulusoys, and Andreas Geiger.
\newblock {Octnet: Learning Deep 3D Representations at High Resolutions}.
\newblock {\em arXiv preprint arXiv:1611.05009}, 2016.

\bibitem{ronneberger2015unet}
Olaf Ronneberger, Philipp Fischer, and Thomas Brox.
\newblock {U-Net: Convolutional Networks for Biomedical Image Segmentation}.
\newblock {\em International Conference on Medical Image Computing and
  Computer-Assisted Intervention}, 2015.

\bibitem{RUMELHART-HINTON-WILLIAMS86}
David~E. Rumelhart, Geoffrey~E. Hinton, and R.~J. Williams.
\newblock Learning internal representations by error propagation.
\newblock In D.~E. Rumelhart, J.~L. McClelland, and the PDP~research group.,
  editors, {\em Parallel distributed processing: Explorations in the
  microstructure of cognition, Volume 1: Foundations}. MIT Press, 1986.

\bibitem{conf/icml/SaxeKCBSN11}
Andrew~M. Saxe, Pang~Wei Koh, Zhenghao Chen, Maneesh Bhand, Bipin Suresh, and
  Andrew~Y. Ng.
\newblock On random weights and unsupervised feature learning.
\newblock In Lise Getoor and Tobias Scheffer, editors, {\em Proceedings of the
  28th International Conference on Machine Learning, ICML 2011, Bellevue,
  Washington, USA, June 28 - July 2, 2011}, pages 1089--1096. Omnipress, 2011.

\bibitem{VGG}
K.~Simonyan and A.~Zisserman.
\newblock Very deep convolutional networks for large-scale image recognition.
\newblock {\em CoRR}, abs/1409.1556, 2014.

\bibitem{OctreeGeneratingNetwork}
Maxim Tatarchenko, Alexey Dosovitskiy, and Thomas Brox.
\newblock {Octree Generating Networks: Efficient Convolutional Architectures
  for High-resolution 3D Outputs}.
\newblock 2017.

\bibitem{journals/corr/TranBFTP14}
Du~Tran, Lubomir Bourdev, Rob Fergus, Lorenzo Torresani, and Manohar Paluri.
\newblock Learning spatiotemporal features with 3d convolutional networks.
\newblock In {\em 2015 IEEE International Conference on Computer Vision
  (ICCV)}, pages 4489--4497. IEEE, 2015.

\bibitem{OcnnNoGroundState}
Peng-Shuai Wang, Yang Liu, Yu-Xiao Guo, Chun-Yu Sun, and Xin Tong.
\newblock {O-CNN}: octree-based convolutional neural networks for {$3$D} shape
  analysis.
\newblock {\em ACM Transactions on Graphics}, 36(4):72:1--72:??, July 2017.

\bibitem{AdaptiveOCNN}
Peng-Shuai Wang, Yang Liu, Yu-Xiao Guo, Chun-Yu Sun, and Xin Tong.
\newblock {Adaptive O-CNN: A Patch-based Deep Representation of 3D Shapes}.
\newblock {\em ACM Transactions on Graphics (SIGGRAPH Asia)}, 37(6), 2018.

\bibitem{House3D}
Yi~Wu, Yuxin Wu, Georgia Gkioxari, and Yuandong Tian.
\newblock Building generalizable agents with a realistic and rich 3d
  environment.
\newblock {\em arXiv preprint arXiv:1801.02209}, 2018.

\bibitem{zeiler10}
Matthew~D. Zeiler, Dilip Krishnan, Graham~W. Taylor, and Rob Fergus.
\newblock {Deconvolutional Networks}.
\newblock {\em Proceedings of the IEEE Conference on Computer Vision and
  Pattern Recognition}, 2010.

\bibitem{AdaptiveDeconvolutional}
Matthew~D. Zeiler, Graham~W. Taylor, and Rob Fergus.
\newblock Adaptive deconvolutional networks for mid and high level feature
  learning.
\newblock In Dimitris~N. Metaxas, Long Quan, Alberto Sanfeliu, and Luc J.~Van
  Gool, editors, {\em IEEE International Conference on Computer Vision, ICCV
  2011, Barcelona, Spain, November 6-13, 2011}, pages 2018--2025. IEEE Computer
  Society, 2011.

\bibitem{CycleGAN}
Jun-Yan Zhu, Taesung Park, Phillip Isola, and Alexei~A Efros.
\newblock Unpaired image-to-image translation using cycle-consistent
  adversarial networks.
\newblock In {\em Computer Vision (ICCV), 2017 IEEE International Conference
  on}, 2017.

\end{thebibliography}

\end{document}